\documentclass[11pt]{article}
\usepackage[margin=1in]{geometry}
\usepackage{multirow}
\usepackage{amsmath}
\usepackage{amssymb}
\usepackage{graphicx}
\usepackage{url}

\usepackage{makecell}
\usepackage{booktabs}
\usepackage{xcolor}
\usepackage{subcaption}

\usepackage[colorlinks=true,linkcolor=red,citecolor=blue,urlcolor=blue]{hyperref}

\makeatletter
\renewcommand{\paragraph}[1]{\textbf{#1}\quad}
\makeatother
\title{Learning Direct Control Policies with Flow Matching for Autonomous Driving}
\date{\textit{Accepted for oral presentation at IEEE ITSC 2026}}

\author{
	\parbox{\textwidth}{%
		\centering
		Marcello Ceresini$^{\dag 1}$ \\ 
        Federico Pirazzoli$^{1}$,
        Andrea Bertogalli$^{1}$,
        Lorenzo Cipelli$^{1}$,
        Filippo D'Addeo$^{2}$ \\
        Anthony Dell'Eva$^{3}$,
        Alessandro Paolo Capasso$^{3}$, 
        Alberto Broggi$^{3}$
	}%
}
\begin{document}
	
	\maketitle
	\thispagestyle{plain}

	\footnotetext[1]{Università degli Studi di Parma}
	\footnotetext[2]{Alma Mater Studiorum - Università di Bologna}
	\footnotetext[3]{VisLab (an Ambarella Inc. company)}
	
	\begin{abstract}
		We present a flow-matching planner for autonomous driving that directly outputs actionable control trajectories defined by acceleration and curvature profiles.
		The model is conditioned on a bird's-eye-view (BEV) raster of the surrounding scene and generates control sequences in a small number of Ordinary Differential Equations (ODE) integration steps, enabling low-latency inference suitable for real-time closed-loop re-planning. 
		We train exclusively on urban scenarios (real urban city streets, intersections and roundabouts of the city of Parma, Italy) collected from a 2D traffic simulator with reactive agents, and evaluate in closed-loop on both in-distribution and markedly out-of-distribution environments, including multi-lane highways and unseen urban scenarios. 
		Our results show that the model generalizes reliably to these unseen conditions, maintaining stable closed-loop control and successfully completing scenarios that differ substantially from the training distribution. We attribute this to the BEV representation, which provides a geometry-centric view of the scene that is inherently less sensitive to distributional shifts, and to the flow-matching formulation, which learns a smooth vector field that degrades gracefully under distribution shift.
		We provide video demonstrations of closed-loop behavior at \url{https://marcelloceresini.github.io/DirectControlFlowMatching}.
	\end{abstract}

	\section{Introduction}
	\label{sec:introduction}
	
	Real-time trajectory generation under multi-modal uncertainty remains a central challenge in autonomous driving. Classical approaches decompose the problem into perception, prediction, and planning modules. While effective in structured settings, these pipelines often struggle to generalize to complex, interactive scenarios and require substantial engineering effort to handle conditions not seen during development~\cite{cuiMultimodalTrajectoryPredictions2019b}.
	
	Deep learning methods have gained significant traction, with generative models emerging as particularly expressive tools for capturing multi-modal future behaviors. Diffusion-based planners~\cite{hoDenoisingDiffusionProbabilistic2020a, jannerPlanningDiffusionFlexible2022a} can produce diverse, realistic trajectories but their iterative denoising incurs considerable computational cost, limiting real-time applicability. Flow matching~\cite{lipmanFlowMatchingGenerative2022} offers an attractive alternative: by directly parameterizing an ODE vector field that transports a simple initial distribution to the data distribution, it achieves comparable generative quality with substantially fewer integration steps and simpler training dynamics~\cite{lipmanFlowMatchingGuide2024a, esserScalingRectifiedFlow2024a}. These properties make flow matching well suited for autonomous driving, where low-latency re-planning is tightly coupled with safety.
	
	A common limitation of data-driven planners is the difficulty of evaluating them under realistic conditions. Open-loop metrics on held-out data do not capture compounding errors, while closed-loop evaluation in simulation with reactive agents provides a more faithful picture of real deployment. Furthermore, most prior work evaluates on the same scenario distribution used for training, leaving open the question of whether learned planners can handle situations outside their training distribution---a critical requirement for real-world driving, where the variety of road geometries, traffic patterns, and agent behaviors is effectively unbounded.
	
	In this work, we adopt a conditional flow-matching~\cite{lipmanFlowMatchingGenerative2022} formulation to generate short-horizon control trajectories conditioned on a BEV scene raster, and investigate the generalization capabilities of this approach through extensive closed-loop evaluation. Our contributions are:
	
	\begin{itemize}
		\item A BEV-conditioned flow-matching architecture that directly outputs actionable controls (acceleration and curvature), designed for real-time closed-loop re-planning with a lightweight iterative vector-field predictor.
		\item A systematic study of out-of-distribution generalization: the model is trained exclusively on urban scenarios and roundabouts, and evaluated in closed-loop on new unseen urban scenarios and multi-lane highways. We show that the learned planner generalizes reliably to these conditions without any fine-tuning.
		\item Extensive closed-loop evaluation in a simulated environment with reactive agents, using safety, progress, and comfort metrics aligned with standard planning benchmarks, accompanied by qualitative trajectory analysis and video demonstrations.
	\end{itemize}
	
	Our architecture is designed around a practical compute--latency trade-off: a BEV-based CNN performs a computationally intensive but one-time encoding of the environment, followed by a lightweight U-Net that is invoked multiple times during ODE integration to produce the final control sequence. This separation enables efficient real-time inference.

	\section{Related Work}
	\label{sec:related_work}
	
	\subsection{Planning in Autonomous Driving}
	Classical autonomous driving systems rely on modular pipelines comprising perception, prediction, and planning components, often using optimization-based planners with hand-crafted cost functions. While effective in many driving scenarios, these approaches require substantial engineering effort and lack generalization to novel situations. As the field matured, large-scale benchmarks~\cite{nuplan, ettingerLargeScaleInteractive2021} enabled data-driven and hybrid methods to emerge, demonstrating the potential of learning-based planners to benefit from data volume and model capacity.
	
	\subsection{Flow Matching for Planning}
	Flow matching~\cite{lipmanFlowMatchingGenerative2022, lipmanFlowMatchingGuide2024a} has recently emerged as a compelling alternative to diffusion models for trajectory generation, offering comparable or superior sample quality~\cite{esserScalingRectifiedFlow2024a} with fundamentally faster inference. Where diffusion-based planners~\cite{hoDenoisingDiffusionProbabilistic2020a, jannerPlanningDiffusionFlexible2022a, DBLP:conf/rss/ChiFDXCBS23} require many denoising steps or truncation strategies~\cite{liaoDiffusionDriveTruncatedDiffusion2025, yangDiffusionESGradientFreePlanning2024a} to achieve real-time operation, flow matching parameterizes a continuous-time ODE vector field that can be integrated in a small number of steps.
	
	Recent work has demonstrated the effectiveness of flow matching for driving. GoalFlow~\cite{xingGoalFlowGoalDrivenFlow2025a} showed that high-quality planning can be achieved in a single forward pass via goal-conditioned flow. FlowDrive~\cite{wangFlowDriveModeratedFlow2025a} showed that careful data balancing combined with lightweight multi-step integration enables flow-based planners to match or exceed state-of-the-art diffusion systems. These results establish flow matching as a promising framework for real-time, safety-critical autonomous planning.
	
	\subsection{Generalization in Learned Planners}
	A key challenge for data-driven planners is generalization beyond the training distribution. Models evaluated exclusively on in-distribution data may exhibit brittle behavior when encountering novel road geometries, unseen traffic patterns, or unfamiliar agent behaviors. While several works have studied distributional robustness in prediction and perception~\cite{gargGPECNetGeneralizablePedestrian2024, cuiMultimodalTrajectoryPredictions2019b}, the question of whether generative planners---and flow-matching planners in particular---can generalize across substantially different scenario types in closed-loop remains relatively underexplored. This work directly addresses this gap by training on a restricted urban distribution and systematically evaluating on held-out scenario categories.

	\section{Simulation Environment}
	\label{sec:simulation}
	
\begin{figure}[t]
		\centering
		\includegraphics[width=\linewidth]{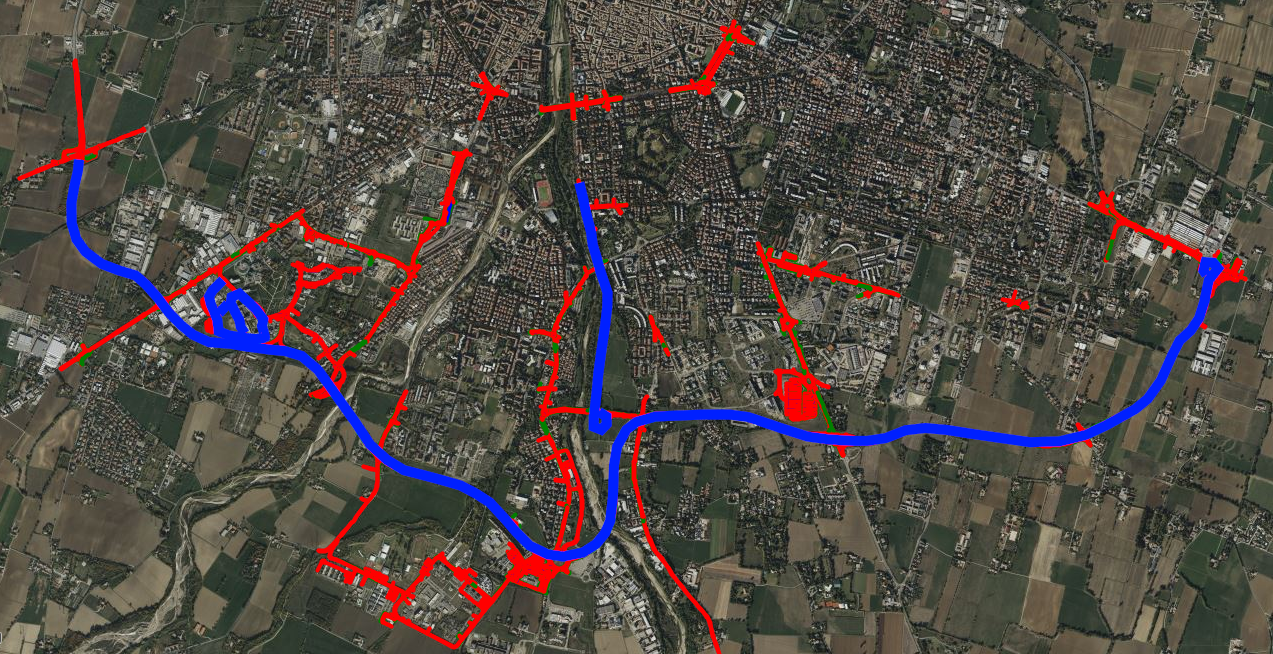}
		\caption{Satellite view of Parma with scenario crop regions. \textcolor{red}{Red} areas denote in-distribution training scenarios (urban streets, roundabouts, intersections), while \textcolor{blue}{blue} areas denote out-of-distribution evaluation scenarios (highway segments and held-out urban configurations).}
		\label{fig:map}
	\end{figure}

	All training and evaluation are conducted in an internally developed 2D traffic simulator that operates at object level on a high-definition (HD) map of the city of Parma, Italy (Figure~\ref{fig:map}). The HD map is derived from OpenStreetMap with manual refinements and encodes lane geometry, lane connectivity, speed limits, drivable area boundaries, stop lines, yield lines, and traffic light positions. All training data, evaluation scenarios, and the HD map will be made publicly available upon publication to enable reproducibility and comparison. Individual scenarios are created by cropping rectangular regions, each capturing a distinct road configuration. In total, we define $50$ real scenarios from the city of Parma spanning a variety of layouts: single- and dual-lane streets, T-intersections, four-way intersections with and without traffic lights, roundabouts of varying sizes, and highway segments with on- and off-ramps.
	
	Of the $50$ scenarios, $40$ are in-distribution urban configurations (single-lane streets, roundabouts, intersections) used for both training data collection and in-distribution evaluation (red regions in Figure~\ref{fig:map}). The remaining $10$ scenarios---highway segments and held-out urban configurations from geographically disjoint map regions---are reserved exclusively for out-of-distribution evaluation (blue regions in Figure~\ref{fig:map}). For closed-loop testing, we define $10$ fixed initial conditions per scenario, yielding $400$ in-distribution and $100$ out-of-distribution evaluation episodes.
	
	\subsection{Simulator Overview}
	
	The simulator supports multi-agent scenarios with heterogeneous traffic participants. At the start of each simulation, all feasible routes between scenario entry and exit points are precomputed, either based on scenario boundaries or defined manually for specific cases. Throughout the simulation, agents are continuously spawned and removed: each agent appears at a designated entry point, follows a route sampled uniformly from the precomputed set, and is removed upon completion. We refer to the agents whose data is recorded for training as \emph{ego agents}, and to all other traffic participants as \emph{non-ego agents}. Multiple ego agents can run in parallel within a single scenario, and multiple scenarios can run simultaneously to speed up data collection. In the closed-loop evaluation setting, a single ego agent is controlled by the learned model while all non-ego agents follow their rule-based policies (Section~\ref{sec:agent_behavior_models}).
	
	We define an \emph{episode} as the interval between an agent's spawn and one of four terminal outcomes: \emph{success} (the agent reaches its assigned destination), \emph{out-of-route} (the agent deviates from its route), \emph{timeout} (the episode exceeds a maximum duration), or \emph{collision}. At each timestep, the ego agent records the poses, speeds, headings, and bounding-box dimensions of all agents in the scene, together with the positions and states of traffic regulations (stop lines, traffic lights). The ego agent also records the route assigned to it at the start of its episode. To ensure traffic diversity, agent-level parameters---initial speed, desired cruising speed, aggressiveness, merging assertiveness, and acceleration limits---are sampled independently for each agent. Traffic density is set per scenario and can vary over the course of a simulation to model different congestion levels. The resulting episodes cover a wide range of driving situations: lane following and adaptive cruise control, roundabout entry and navigation, highway merging and lane changes, intersection handling (with and without traffic lights), and avoidance of static or dynamic obstacles on the road.
	
	For training data, we retain only successful ego-agent episodes, ensuring that the model is trained exclusively on correct demonstrations of driving behavior. The resulting training set consists of ${\sim}19\text{K}$ successful episodes, corresponding to approximately $35$ hours of simulated driving. Although this may appear modest compared to large-scale driving datasets, our data is collected at $20\,\mathrm{Hz}$ re-planning frequency---substantially higher than typical benchmarks---and is actively curated for diversity: we discretize the state space of vehicle acceleration and curvature into bins and selectively store episodes that populate underrepresented regions, continuously monitoring the distribution during data collection. This ensures that the dataset is dense in informative driving situations rather than dominated by trivial straight-line driving.
		
	As a 2D object-level simulator, the environment does not model sensor noise, occlusions, or localization errors. These simplifications are partially mitigated by the data augmentation applied dynamically over the raw data during training (Section~\ref{sec:dataset}), and we discuss remaining limitations in Section~\ref{sec:discussion}.
	
	\subsection{Agent Behavior Models}
	\label{sec:agent_behavior_models}
	
	The simulator supports several categories of traffic participants:
	
	\begin{itemize}
		\item \textbf{Vehicles} (cars, trucks, buses, motorcycles): all vehicle agents share the same planning architecture, consisting of an Intelligent Driver Model~\cite{IDM_controller} (IDM) for longitudinal control and a Pure Pursuit~\cite{PurePursuit_controller} controller for lateral path tracking. Each agent is parameterized with a configurable aggressiveness profile that modulates desired speed, following distance, merging assertiveness, and acceleration limits. Vehicle dynamics are governed by a kinematic bicycle model.
		\item \textbf{Pedestrians and cyclists:} these agents follow a predefined route along the road boundary and may cross the road at crosswalks or random locations, traveling in the same or opposite direction as traffic.
		\item \textbf{Vehicle queues:} sequences of stationary or slow-moving vehicles that model congestion at intersections or traffic lights.
		\item \textbf{Static obstacles:} generic objects placed at the roadside or within the driving lane to simulate parked vehicles or unexpected/undefined obstructions.
	\end{itemize}
	
\begin{figure}[t]
		\centering
		\includegraphics[width=\linewidth]{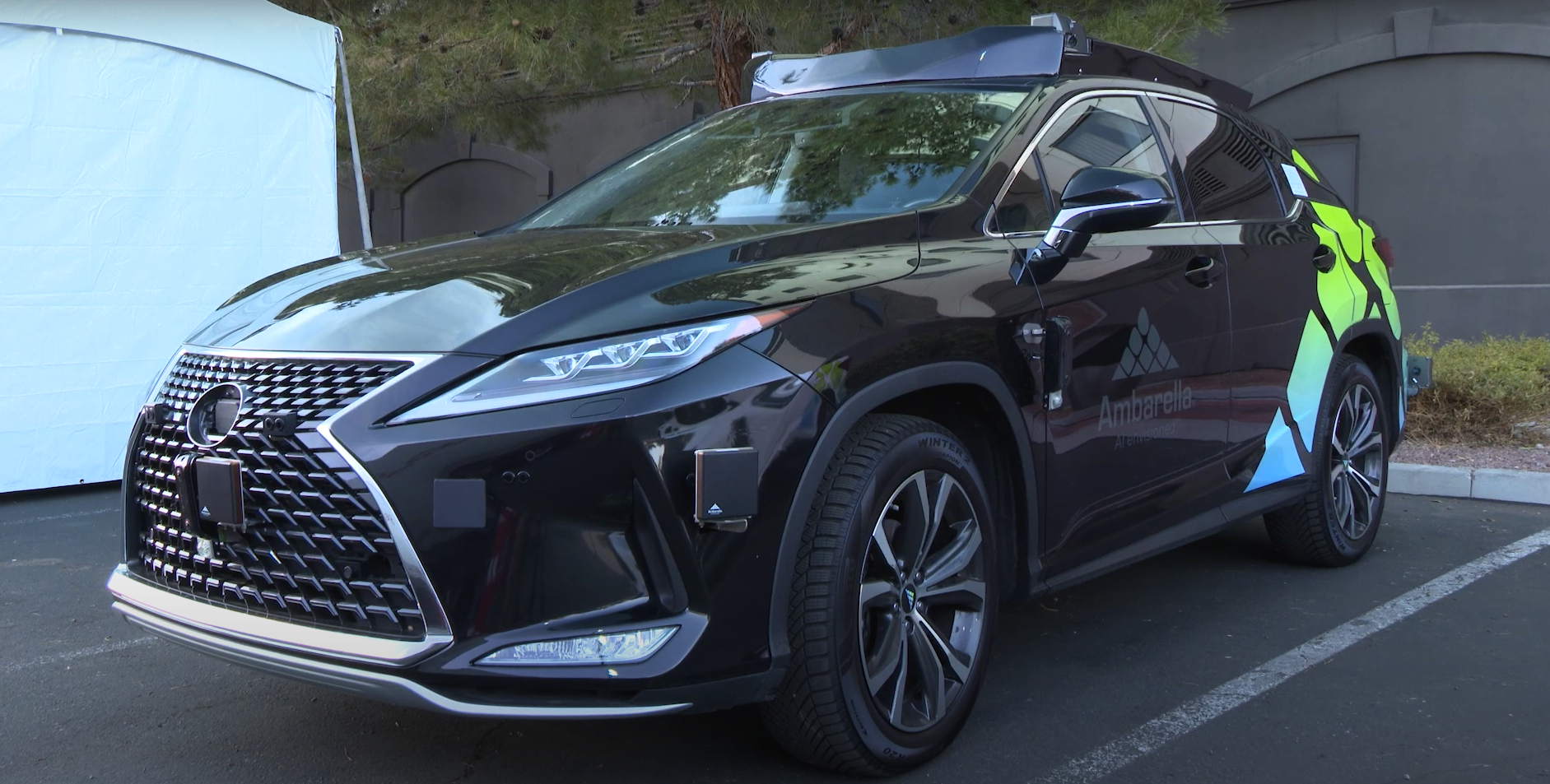}
		\caption{Lexus RX 350h, the autonomous vehicle used for real-world testing. The vehicle is equipped with a variety of sensors, including cameras, radars, and GPS.}
		\label{fig:lexus}
	\end{figure}

	\paragraph{Data collection.}
	During data collection, all agents---including the ego---have access to the full state of the environment as well as the future trajectories of other actors. This privileged information enables near-optimal behaviors, particularly in scenarios that require negotiation among agents (e.g., merging, intersection handling). Consequently, we can generate clean and consistent driving data from the ego agents, which are later used to train the planning model. To produce training data that is representative of real vehicle behavior, the ego agent's dynamics are simulated using a learned deep dynamic model as implemented in~\cite{capassoSimulationRealWorld2020} that approximates the response of our real testing car (Lexus RX 350h, Fig.~\ref{fig:lexus}), facilitating subsequent deployment on hardware.
	
	\paragraph{Closed-loop evaluation.}
	In the closed-loop setting, the ego agent no longer has access to the future routes of surrounding actors. Instead, it must implicitly infer their behavior from its observation of the environment (see Section~\ref{sec:dataset}) and predict the control action that best satisfies the driving objective, closely reflecting the conditions encountered during real-world deployment. All non-ego agents always remain \emph{reactive}: their actions are recomputed at every timestep in response to the ego's actual driven trajectory (and the other agents in the scene), rather than replayed from pre-recorded data.

	\section{Dataset}
	\label{sec:dataset}
	
	\subsection{Data Collection}
	
	Training data is collected by rolling out the rule-based agent policies described in Section~\ref{sec:simulation} within the simulator. During the simulation, the ego-agent (running at 20Hz) records at each timestep the following information:
	\begin{itemize}
		\item The \textbf{ego-agent target action} $(a, \kappa)$, i.e.\ the acceleration and curvature command that is passed through the dynamic model to update the ego pose. This serves as the ground-truth supervision signal.
		\item The \textbf{poses and speeds} of all agents in the scene (ego and non-ego), together with their object class (car, bus, bicycle, pedestrian, etc.), bounding-box dimensions (width, length), and heading.
	\end{itemize}
	
	Additionally, we record the route assigned to the ego agent at the start of each episode.
	The training set comprises $2{,}524{,}298$ frames collected exclusively from urban and roundabout scenarios. No separate held-out test set is used: all evaluation is performed in closed-loop (Section~\ref{sec:closed_loop_eval}), where the model drives the ego vehicle in real time within the simulator. Table~\ref{tab:dataset_stats} summarizes the training data composition.
	
	\subsection{BEV Representation}
	
\begin{figure*}[ht]
		\centering
		\begin{subfigure}{0.24\linewidth}
			\centering
			\includegraphics[width=\linewidth]{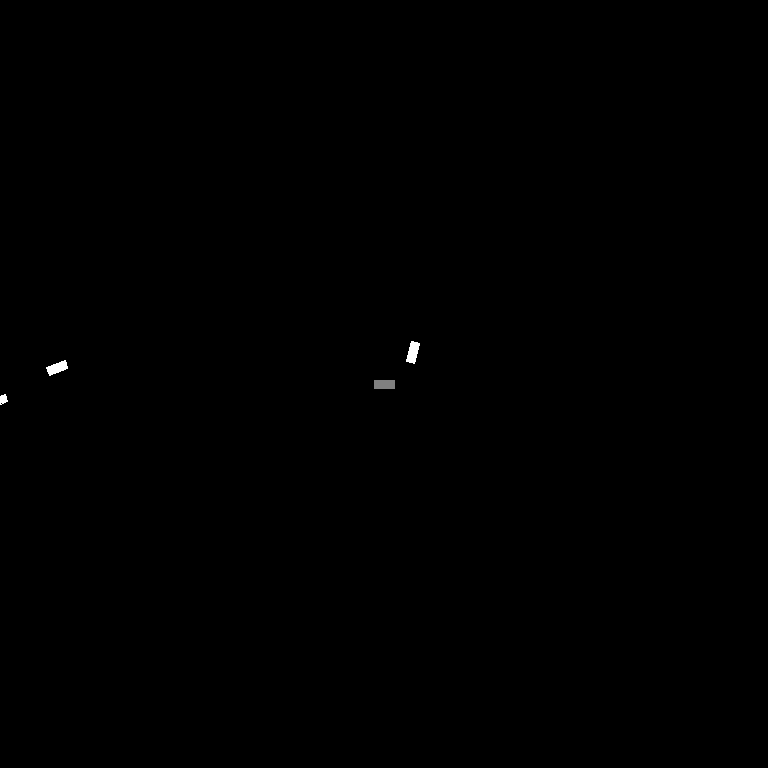}
			\subcaption{Obstacles}
		\end{subfigure}\hfill
		\begin{subfigure}{0.24\linewidth}
			\centering
			\includegraphics[width=\linewidth]{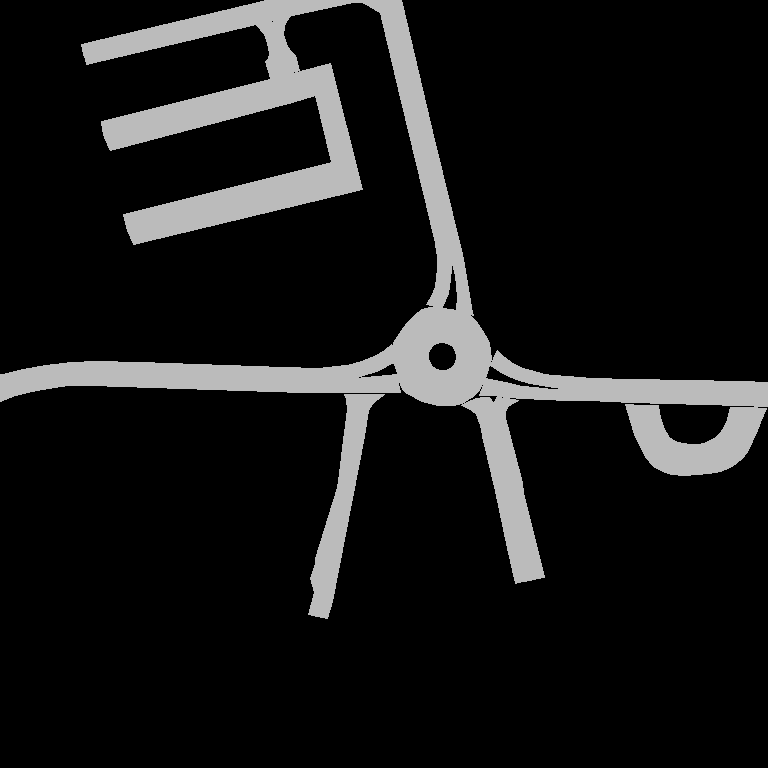}
			\subcaption{Drivable area}
		\end{subfigure}\hfill
		\begin{subfigure}{0.24\linewidth}
			\centering
			\includegraphics[width=\linewidth]{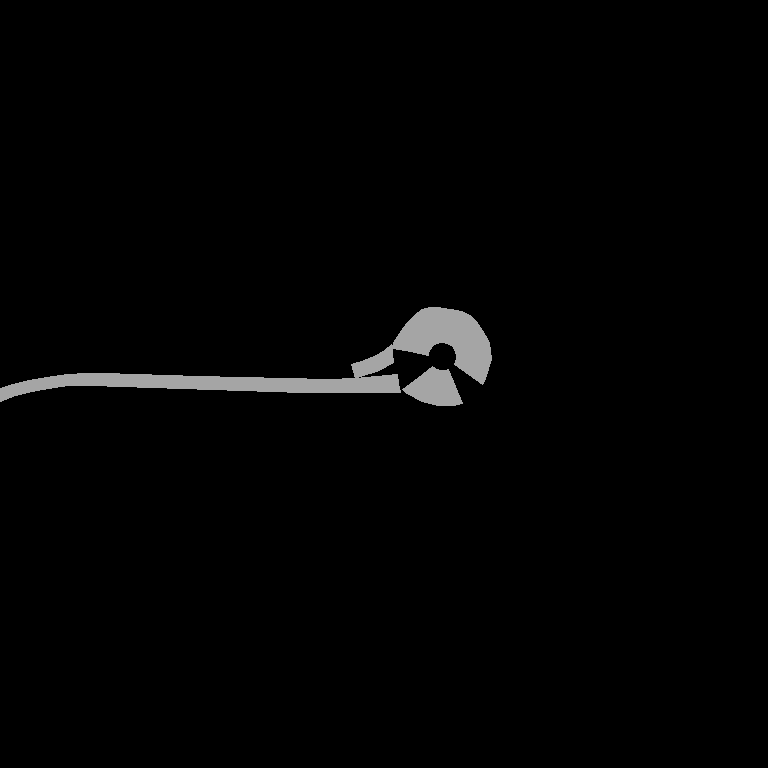}
			\subcaption{Ego route}
		\end{subfigure}\hfill
		\begin{subfigure}{0.24\linewidth}
			\centering
			\includegraphics[width=\linewidth]{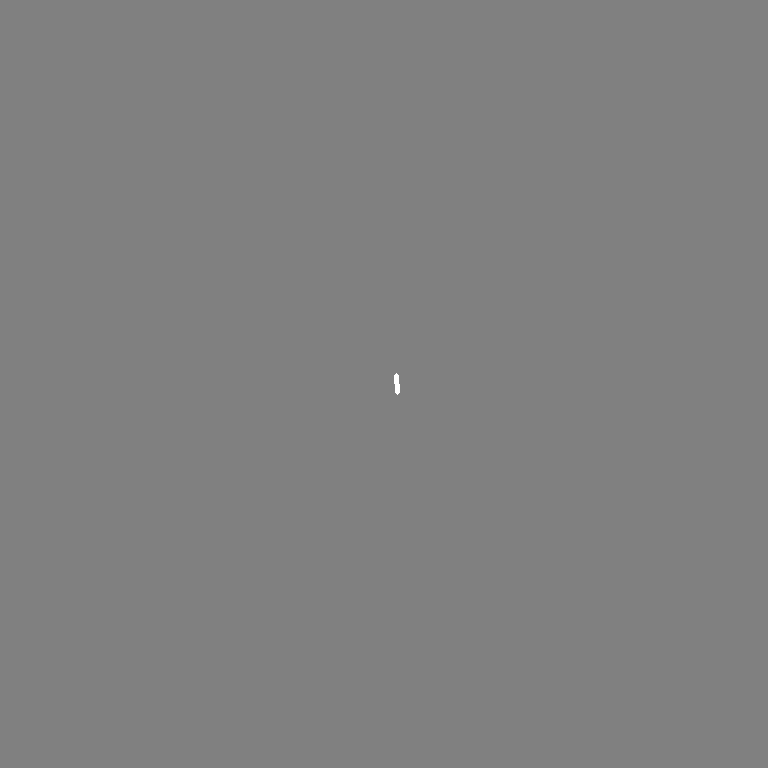}
			\subcaption{Regulations}
		\end{subfigure}
		\caption{Visualization of the four BEV channels: (a) obstacles, (b) drivable area, (c) ego route, and (d) regulations.}
		\label{fig:bev_channels}
	\end{figure*}

	The vectorized scene information recorded during data collection---agent poses, speeds, bounding-box dimensions, object classes, ego-agent route and HD map geometry---is rasterized into a bird's-eye-view (BEV) tensor at training time. Each training sample is therefore a tuple $(\mathbf{B}, z)$ where $\mathbf{B} \in \mathbb{R}^{4 \times h \times w}$ is the BEV raster and $z \in \mathbb{R}^{n \times 2}$ is the ground-truth control sequence. The BEV raster has spatial resolution $h = w = 768$ at $0.25\,\mathrm{m}$ per pixel, covering a $192\,\mathrm{m} \times 192\,\mathrm{m}$ area centered on the ego agent ($96\,\mathrm{m}$ in each direction). The four channels (Figure~\ref{fig:bev_channels}) encode:
	
	\begin{enumerate}
		\item \textbf{Obstacles:} The ego-agent's bounding box is rendered with a constant value; other agents are rasterized with pixel values proportional to their speed.
		\item \textbf{Drivable area:} Drivable road surface with pixel values encoding the speed limit.
		\item \textbf{Ego route:} The planned route rendered as the union of drivable lanes that compose the ego-agent's route, with pixel values encoding the ego's current speed.
		\item \textbf{Regulations:} A single channel indicating stop locations (yield lines, stop signs, traffic lights) with different pixel values for different types of regulations.
	\end{enumerate}
	
	To improve robustness to the imperfect perception that would be encountered in a real-world system, we apply data augmentation during training by randomly perturbing the poses and speeds of non-ego agents in the BEV raster. This injects noise that approximates real-world perception errors and encourages the model to learn policies that are less sensitive to small inaccuracies in the observed scene. On our autonomous vehicle the BEV representation is based on data retrieved from perception and localization modules. Future work will aim at fusing these modules for real-world evaluation of the model.
	
	\subsection{Control Representation}
	
	The ground-truth control sequence $z \in \mathbb{R}^{n \times 2}$ consists of $n = 64$ future timesteps at $20\,\mathrm{Hz}$ (a horizon of $3.2\,\mathrm{s}$). The two control dimensions are acceleration $a \in [-3, 2]\,\mathrm{m/s^2}$ and curvature $\kappa = 1/r \in [-0.2, 0.2]\,\mathrm{m^{-1}}$, where $r$ is the turning radius. Both are normalized before being fed to the model.
	
	\subsection{Dataset Statistics and Distribution}
	
\begin{figure}[h]
	  \centering
	  \includegraphics[width=\linewidth]{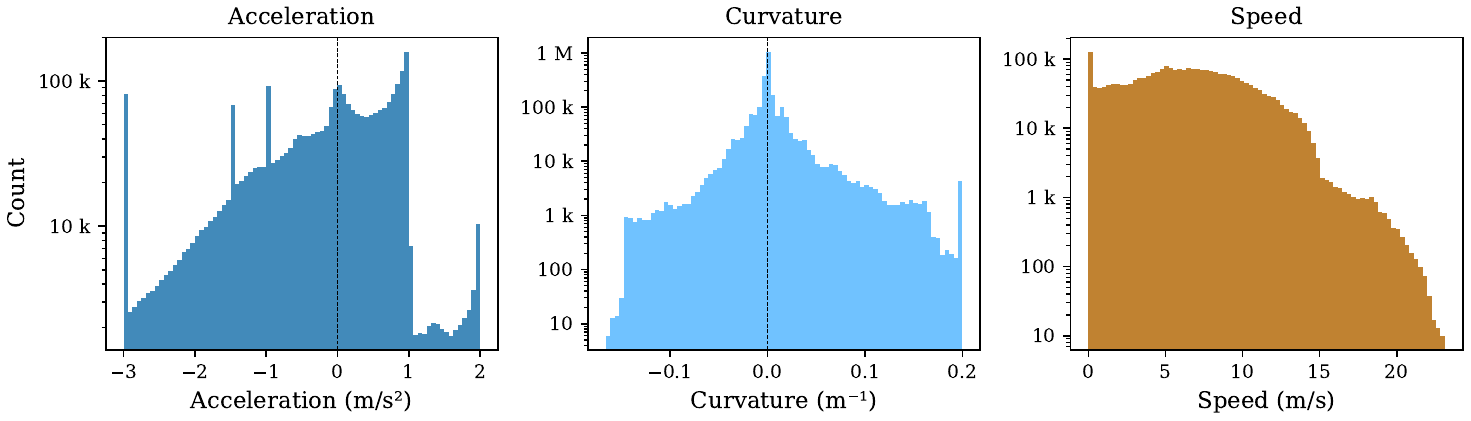}
	  \caption{Distribution of ego-agent controls and speed in the training set.
	           Left: acceleration. Center: curvature. Right: speed.}
	  \label{fig:dataset_histograms}
	\end{figure}

	Understanding the composition of the training data is important for interpreting the generalization results. Table~\ref{tab:dataset_stats} summarizes the key statistics and Figure~\ref{fig:dataset_histograms} shows control and speed distributions.
	
	\begin{table}[h]
	\caption{Training dataset statistics.}
	\label{tab:dataset_stats}
	\centering
	\begin{tabular}{l r}
	\toprule
	\textbf{Statistic} & \textbf{Value} \\
	\midrule
	Total training frames      & 2,524,298 \\
	\quad Urban                & 1,573,183 (62.3\%) \\
	\quad Roundabouts          & 951,115 (37.7\%) \\
	\midrule
	Ego speed                  & $6.36 \pm 3.71$  m/s \\
	Mean acceleration          & $-0.177 \pm 1.012$ m/s$^2$ \\
	Mean $|\kappa|$            & $0.0026 \pm 0.0269$ m$^{-1}$ \\
	\% frames stopped          & 4.5\% \\
	\bottomrule
	\end{tabular}
	\end{table}
	
	The training distribution is dominated by straight driving and gentle acceleration, with sharp steering and strong braking events being substantially rarer. This imbalance reflects realistic urban driving conditions, where the majority of time is spent in steady-state lane following, but can be easily rebalanced with a custom data sampler to enhance the importance of rarer control sequences. The curvature distribution is concentrated near zero, with a long tail corresponding to roundabout and turn maneuvers.
	
	
	\section{Model}
	\label{sec:model}
	
	\subsection{Conditional Flow Matching}
	
	Flow matching~\cite{lipmanFlowMatchingGenerative2022} provides a framework for transforming a simple initial distribution $p_\text{init}$ into a data distribution $p_\text{data}$ via an ODE. A trajectory $X: [0,1] \to \mathbb{R}^d$ is defined as the solution of
	\begin{equation}
		\frac{d}{dt} X_t = u_t^\theta(X_t), \quad X_0 \sim p_\text{init},
		\label{eq:ode}
	\end{equation}
	where $u_t^\theta$ is a learned vector field. Sampling reduces to drawing $X_0 \sim p_\text{init}$ and integrating numerically (e.g.\ Euler: $X_{t+h} = X_t + h\, u_t^\theta(X_t)$).
	
	Following~\cite{lipmanFlowMatchingGenerative2022, esserScalingRectifiedFlow2024a}, we choose a linear interpolation schedule $\alpha_t = t$, $\beta_t = 1 - t$, yielding the interpolated state $x_t = t z + (1-t) \epsilon$ with $z \sim p_\text{data}$ and $\epsilon \sim \mathcal{N}(0, I)$. The conditional flow matching objective simplifies to:
	\begin{equation}
		\mathcal{L}_\text{CFM}(\theta) = \mathbb{E}_{t, z, \epsilon} \left\| u_t^\theta(x_t) - (z - \epsilon) \right\|_2^2,
		\label{eq:cfm_loss}
	\end{equation}
	where $t \sim \mathcal{U}(0,1)$. The training target $z - \epsilon$ is independent of $t$, producing straight-line trajectories between noise and data, which reduces the number of integration steps needed for accurate sampling.
	
	In our setting, the data distribution corresponds to ground-truth control sequences $z \in \mathbb{R}^{n \times 2}$ (acceleration and curvature), and the model generates control sequences by integrating Eq.~\eqref{eq:ode} from an initial state $X_0$. For reproducibility, we initialize all inference from the mean of the Gaussian prior, i.e.\ $X_0 = \mathbf{0}$, rather than sampling stochastically. This yields deterministic outputs for a given BEV input.
	
	\subsection{Architecture}
	
\begin{figure*}[ht]
		\centering
		\includegraphics[width=\linewidth]{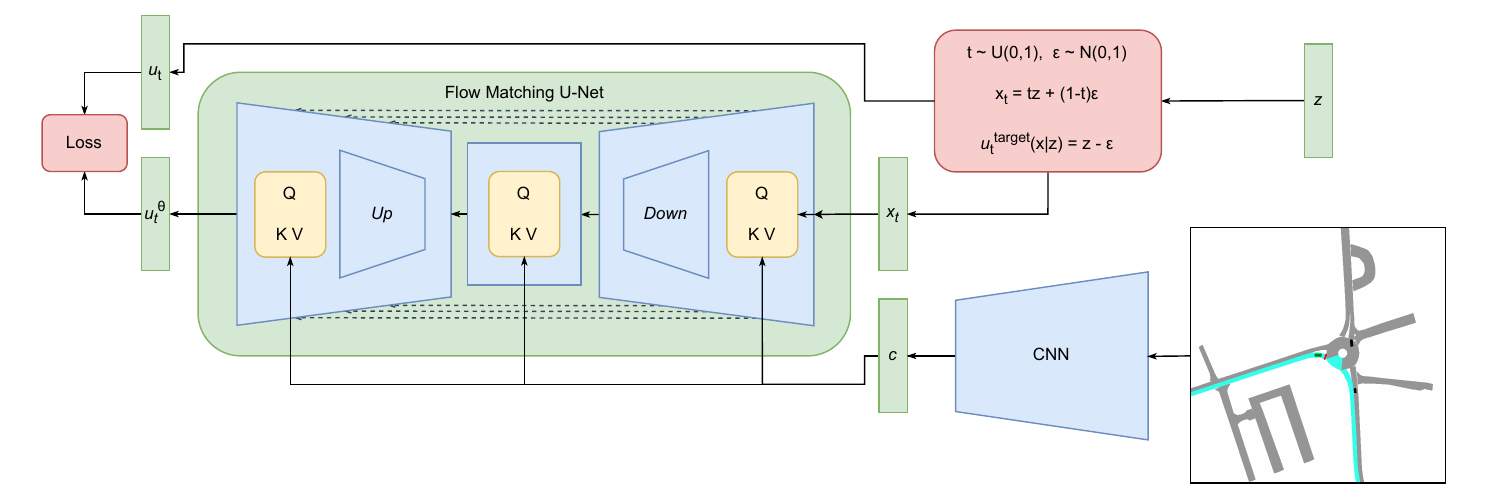}
		\caption{Overview of the BEV-conditioned flow-matching architecture. The BEV raster is encoded once by a CNN, and the resulting embedding is injected via cross-attention into a lightweight U-Net that iteratively refines the control trajectory through ODE integration.}
		\label{fig:conditioning}
	\end{figure*}

	Since our goal is real-time closed-loop deployment, we design the architecture around a separation of concerns: a heavy one-time encoding of the scene, followed by a lightweight iterative generator (Figure~\ref{fig:conditioning}).
	
	\paragraph{BEV encoder.}
	The BEV raster $\mathbf{B} \in \mathbb{R}^{4 \times h \times w}$ is processed by a CNN to produce a compact spatial embedding $c = \text{CNN}_\text{BEV}(\mathbf{B}) \in \mathbb{R}^m$. This encoding is computed once per planning cycle and reused across all ODE integration steps.
	
	\paragraph{Vector-field U-Net.}
	The vector field $u_t^\theta(\cdot \mid c)$ is parameterized by a U-Net~\cite{ronnebergerUNetConvolutionalNetworks2015} that takes the noisy control sequence $x_t$ and the interpolation time $t$ as input, and is conditioned on the BEV embedding $c$ via cross-attention, following the conditioning paradigm of~\cite{rombachHighResolutionImageSynthesis2022a}. The U-Net consists of two down-blocks, a middle block, and two up-blocks, with cross-attention layers placed at the skip connections and in the mid-block for effective multi-scale fusion. The full model contains approximately $15$M parameters, of which roughly $85\%$ reside in the BEV CNN. The U-Net is deliberately kept lightweight because it is executed at every ODE integration step during inference.
	
	\subsection{Training}
	
	The BEV encoder and the vector-field U-Net are trained jointly end-to-end from scratch with the CFM objective (Eq.~\eqref{eq:cfm_loss}). We use AdamW~\cite{loshchilovDecoupledWeightDecay2018} ($\beta_1 = 0.9$, $\beta_2 = 0.999$, weight decay $0.1$) at a peak learning rate of $5 \times 10^{-4}$. Training runs for $300$k steps with a batch size of $32$, using a linear warmup over the first $5\%$ of steps followed by cosine annealing. On a single NVIDIA RTX A6000~\cite{nvidia2020rtxa6000}, training completes in approximately three days.

	\section{Closed-Loop Evaluation}
	\label{sec:closed_loop_eval}
	
	\subsection{Closed-Loop Protocol}
	
	At each ego-agent timestep ($20\,\mathrm{Hz}$), the evaluation loop proceeds as follows: (i) a BEV raster is rendered from the current ego state, all other agent states, the local HD map geometry and the ego-agent route; (ii) the flow-matching model generates a control sequence by integrating the learned ODE from the zero initial state; (iii) only the first predicted control $(a_1, \kappa_1)$ is applied to update the ego vehicle's state via the deep dynamic model; (iv) all non-ego agents update their states according to their reactive rule-based policies (Section~\ref{sec:simulation}); (v) the process repeats from~(i).
	
	Unlike replay-based evaluation, where non-ego agents follow pre-recorded trajectories regardless of the ego's behavior, non-ego agents re-plan at every timestep in response to the ego's actual driven trajectory, as explained in Section~\ref{sec:simulation}. This makes the evaluation substantially more challenging and realistic, as the ego must handle situations that emerge from the interaction between its own actions and the adaptive responses of other traffic participants.
	
	In contrast to the data-collection phase, where scenario parameters (starting positions, speeds, aggressiveness profiles) are sampled randomly, evaluation episodes use fixed initial conditions. This ensures that each scenario is fully reproducible in every evaluation run.
	
	\subsection{In-Distribution vs.\ Out-of-Distribution Scenarios}
	
	We evaluate the model in two regimes. \emph{In-distribution} episodes use scenario types present in the training data---urban streets, intersections, and roundabouts---but with fixed initial conditions. \emph{Out-of-distribution} (OOD) episodes use scenario types never seen during training, drawn from geographically disjoint map regions (see Figure~\ref{fig:map}) to eliminate any risk of train--test leakage.
	
	The primary source of distribution shift in the OOD set is the inclusion of highway scenarios. These differ from the training distribution in several important ways: (i) target speeds for both ego and non-ego agents are substantially higher (up to $90\,\mathrm{km/h}$ vs.\ the urban speeds seen during training), so that speed-encoded pixel values in the BEV raster occupy a region never encountered during training; (ii) the vehicle dynamics at highway speeds---longer braking distances, different steering sensitivity---have never been experienced by the model; and (iii) the combination of high speeds and a forward field of view limited to $96\,\mathrm{m}$ (half the $192\,\mathrm{m}$ BEV extent) means that oncoming agents appear and close in much more quickly, possibly requiring faster reactions with less planning horizon.
	
	\subsection{Metrics}
	\label{subsec:metrics}
	
	We adopt metrics aligned with established closed-loop planning benchmarks such as nuPlan~\cite{nuplan} and NAVSIM~\cite{navsim}, adapted to our simulation setup. All metrics are computed on driven trajectories produced by the model in closed-loop and reported in Table~\ref{tab:cl_comparison}.
	
	\begin{itemize}
		\item \textbf{Collision rate (CR):} fraction of episodes in which the ego vehicle collides with any other agent or static obstacle.
		\item \textbf{Drivable area compliance (DAC):} fraction of episodes in which the ego vehicle remains within the drivable road surface at all times. An episode is marked as non-compliant if any corner of the ego bounding box exits the drivable area.
		\item \textbf{Route progress (RP):} fraction of the planned route completed by the ego vehicle, averaged across episodes. This captures partial success even when the ego does not reach the final goal.
	\end{itemize}
	
	To evaluate ride quality, we define two complementary jerk-based metrics computed directly from the predicted controls. Let $f = 20\,\mathrm{Hz}$ denote the simulation and re-planning framerate.
	\begin{itemize}
		\item \textbf{Predicted-sequence jerk ($J_\text{seq}$):} at each re-planning step, the model outputs a sequence of $n$ future controls. We compute the longitudinal jerk as the finite difference of successive accelerations, $j_i = (a_{i+1} - a_i) \cdot f$, and report the mean absolute jerk across all timesteps and episodes. This measures the smoothness of the \emph{planned} control profile.
		\item \textbf{Executed jerk ($J_\text{exec}$):} across successive re-planning steps, only the first control $(a_1^{(t)}, \kappa_1^{(t)})$ is actually applied. We compute the longitudinal jerk of the \emph{executed} trajectory as $j^{(t)} = (a_1^{(t+1)} - a_1^{(t)}) \cdot f$ and report the mean absolute value. This captures the smoothness of the actual driving behavior as experienced by passengers, including inconsistencies between successive re-planning outputs.
	\end{itemize}
	
	Only for qualitative evaluation, we apply a bicycle model~\cite{rajamaniVehicleDynamicsControl2012} to convert the predicted controls into vehicle poses, which are used to visualize predicted future trajectories in the supplementary videos.

	\section{Results}
	\label{sec:results}
	
	\subsection{Integration Steps and Solver Selection}
	
	A key practical consideration for flow-matching planners is the number of function evaluations (NFE) used during ODE integration at inference time. Each NFE requires one forward pass through the vector-field U-Net, so the NFE directly determines the latency of control generation. We compare several ODE solvers across different NFE values.
	
\begin{figure}[h]
		\centering
		\includegraphics[width=0.9\linewidth]{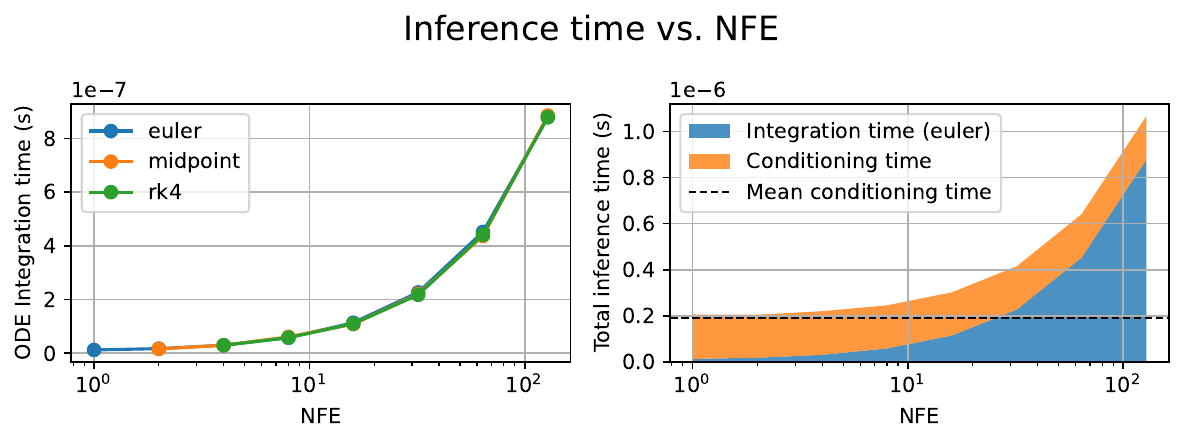}
		\caption{Left: ODE integration time vs.\ NFE by solver. Right: fraction of total inference time spent on ODE integration as NFE increases.}
		\label{fig:inference_time}
	\end{figure}

	We observed no meaningful difference in closed-loop behavior between the Euler method and higher-order solvers (Heun, RK4), so we adopt Euler for its simplicity. At $\text{NFE}=10$ the total forward computation time is around $25\,\mathrm{ms}$ (Figure~\ref{fig:inference_time}), which is well within the real-time latency budget that would be required for closed-loop re-planning on an NVIDIA RTX A6000 GPU.
	
	Based on this analysis, we select the Euler solver with $\text{NFE}=10$ as our primary configuration for all closed-loop experiments. We additionally report results with $\text{NFE}=1$ (single-step generation) in Table~\ref{tab:cl_comparison} to quantify the impact of the number of integration steps on closed-loop performance.
	
	\subsection{Closed-Loop Performance}
	
	\begin{table}[h]
	\caption{Closed-loop results: in-distribution vs.\ out-of-distribution. NFE indicates the number of ODE integration steps used by the flow-matching model.}
	\label{tab:cl_comparison}
	\centering
	\begin{tabular}{l cc cc}
	\toprule
	& \multicolumn{2}{c}{\textbf{NFE=1}} & \multicolumn{2}{c}{\textbf{NFE=10}} \\
	\cmidrule(lr){2-3} \cmidrule(lr){4-5}
	\textbf{Metric} & ID & OOD & ID & OOD \\
	\midrule

	CR (\%) $\downarrow$                   & 9.25\% & 7\% & 3.75\% & 2.75\% \\
	DAC (\%) $\uparrow$                    & 82\% & 96\% & 93.75\% & 98\% \\
	RP (\%) $\uparrow$                     & 92.37\% & 93.51\% & 98.12\% & 98.32\% \\
	$J_\text{seq}$ (m/s$^3$) $\downarrow$  & 0.12 & 0.09 & 0.07 & 0.06 \\
	$J_\text{exec}$ (m/s$^3$) $\downarrow$ & 2.13 & 1.65 & 1.32 & 1.19 \\
	\bottomrule
	\end{tabular}
	\end{table}
	
	Table~\ref{tab:cl_comparison} reports closed-loop metrics for both NFE configurations across in-distribution and OOD episodes. With $\text{NFE}=10$, the model achieves low collision rates, high drivable area compliance, and near-complete route progress in both regimes. Increasing from $\text{NFE}=1$ to $\text{NFE}=10$ yields consistent improvements across all metrics, confirming that the performance increase of multi-step ODE integration translates to improved closed-loop control quality.
	
	A notable observation is that OOD performance is comparable to---and in some metrics slightly better than---in-distribution performance. This is because highway scenarios involve fewer complex decision points (merges, tight turns, intersection negotiations) that require precise control, and the lower density of close-range interactions reduces the probability of collisions caused by aggressive non-ego agents. However, as discussed below, this quantitative advantage masks qualitative differences in driving behavior at highway speeds.
	
	The comfort metrics reveal a striking gap between the predicted-sequence jerk $J_\text{seq}$ and the executed jerk $J_\text{exec}$: $J_\text{seq}$ is nearly two orders of magnitude lower. This indicates that the model produces smooth, internally consistent control sequences at each re-planning step, but exhibits higher sensitivity to frame-to-frame variations in the BEV input, leading to larger differences between successive first-step controls. Addressing this gap---for instance by conditioning on temporal sequences of BEV rasters rather than single frames---is an interesting direction for future work.
	
	\subsection{Qualitative Analysis}
	
	
	\paragraph{In-distribution scenarios.}
	The model reliably follows the assigned route across all in-distribution scenario types, including lane following, adaptive cruise control (following a lead vehicle), left and right turns, and roundabout entry and exit. In standard driving situations the behavior is smooth and closely resembles the expert demonstrations. The $J_\text{exec}$ metric is in all settings below $2.5$ m/s$^3$, which is a good indication of smooth driving behavior and below commonly accepted thresholds for ride comfort~\cite{nuplan}.
	
	The main difficulty arises in tight maneuvers: on sharp turns, the model tends to cut the corner slightly and then overshoot, suggesting that it cannot reach the high curvature values needed for these situations. This is consistent with the training distribution, where high-curvature frames are slightly underrepresented (Section~\ref{sec:dataset}).
	
	The occasional drivable-area violations visible in the DAC metric can be attributed to the simulator's representation of road boundaries: there is no explicit distinction between lane markings and physical curbs, so boundaries are effectively ``soft.'' The expert planner used for data collection treats them as such---occasionally cutting corners, particularly in tight turns---and the learned model reproduces this behavior. This is compounded by the well-known tendency of the Pure Pursuit controller (used in data collection) to cut trajectories on curved paths. In future work, we will investigate the impact of different data-collection deterministic controllers on the model's behavior.
	
	\paragraph{Out-of-distribution scenarios.}
	On highway scenarios, the model successfully maintains lane following and adapts its speed to surrounding traffic despite never having seen these speeds or road geometries during training. Quantitatively, OOD metrics are slightly better than in-distribution, mainly because highway driving involves fewer complex interactions and less exposure to aggressive non-ego behavior. Also the jerk metrics are slightly lower on OOD episodes, which can be attributed to the lower variation of road geometry (and in turn, BEV raster inputs) during the episode, with highways consisting of long straight lines or low and constant curvature long turns.
	
	However, the supplementary videos reveal qualitative differences: at highway speeds the ego vehicle struggles more to maintain precise lane centering, likely due to the unfamiliar vehicle dynamics at high speed and the higher sensitivity of lateral position to small curvature errors. On the other hand, the absence of sharp turns in highway scenarios means the model never drives out of the drivable area, explaining the higher DAC on OOD episodes.
	
	\subsection{Failure Analysis}
	\label{sec:failure}
	
	The dominant failure mode across both in-distribution and OOD episodes is \emph{aggressive non-ego behavior}. This manifests in two ways: non-ego agents spawned immediately behind the ego with a large speed differential whose aggressive driving policy does not brake strongly enough to avoid a rear-end collision, and non-ego agents that cut into the ego's lane during merging or turning maneuvers with an insufficient gap. In both cases, the collision is largely unavoidable regardless of the ego's actions. Because our planner is purely frame-based---it observes a single BEV snapshot with no temporal history---it cannot anticipate such maneuvers long before they materialize, at which point it is often too late to react. A planner with temporal context or explicit agent-intent modeling could mitigate some of these failures, but this falls outside the scope of the present work.

	\section{Discussion and Limitations}
	\label{sec:discussion}
	
	\paragraph{Generalization properties.}
	Our results indicate that a flow-matching planner trained on a restricted urban distribution can generalize to substantially different driving conditions in closed-loop. We hypothesize that this arises from two factors. First, the BEV representation abstracts away scenario-specific visual details and presents the model with a structured, geometry-centric view of the scene, allowing it to learn spatial relationships (e.g., ``slow down near obstacles,'' ``follow the drivable area'') that transfer across road types. Second, the flow-matching objective, by learning a smooth vector field over the space of control sequences, may produce robust representations that only slightly degrade when the input distribution shifts, rather than failing catastrophically. We note that all results are obtained in a 2D object-level simulator; validating the sim-to-real transfer of these findings remains an important direction for future work.

	\paragraph{Near-misses.}
	As discussed in Section~\ref{sec:failure}, the main source of collisions is aggressive non-ego behavior that leaves the ego no feasible avoidance strategy. A promising direction to reduce this failure mode is to improve the deterministic planner used for data collection so that it explicitly handles near-miss situations---e.g., by braking defensively or yielding preemptively when an aggressive agent approaches. This would increase the proportion of successful episodes in the training data even under adversarial non-ego conditions, providing the learned model with demonstrations of robust defensive driving and likely improving its collision avoidance in closed-loop evaluation.

	\paragraph{Interactive agents.}
	The rule-based agents used in our simulator, while reactive, follow relatively simple policies. Real-world traffic involves more diverse and less predictable behaviors. It remains to be seen whether the generalization we observe extends to environments with more complex agent interactions, such as aggressive merging, jaywalking pedestrians, or adversarial behavior.

	\section{Conclusion}
	\label{sec:conclusion}
	
	We presented a conditional flow-matching planner that generates control trajectories for autonomous driving from BEV scene rasters. The model produces actionable acceleration and curvature sequences in real time via lightweight ODE integration, and is trained end-to-end with the standard flow-matching objective.
	
	Our central finding is that this approach generalizes beyond its training distribution: trained exclusively on urban scenarios and roundabouts, the model successfully navigates multi-lane highways and unseen urban scenarios in closed-loop simulation with reactive agents, while maintaining low predicted-sequence jerk that indicates smooth and comfortable planned control profiles.
	
	We see several directions for future work. First, incorporating richer scene representations---such as temporal BEV sequences or vectorized map encodings---may further improve the model's ability to reason about dynamic interactions. Second, evaluating sim-to-real transfer, particularly the impact of noisy or imperfect BEV inputs, is essential for understanding real-world applicability. Finally, integrating the approach with established public benchmarks would enable direct comparison with existing planners and strengthen the empirical foundation of the generalization findings reported here.

	\bibliographystyle{plain}
	\bibliography{bibliography} 
	
\end{document}